\documentclass[acmsmall, nonacm]{acmart}
\usepackage{algorithm2e}
\AtBeginDocument{%
  \providecommand\BibTeX{{%
    \normalfont B\kern-0.5em{\scshape i\kern-0.25em b}\kern-0.8em\TeX}}}

\begin{document}

%%
%% The "title" command has an optional parameter,
%% allowing the author to define a "short title" to be used in page headers.
\title{A Survey of the Various Methodologies Towards making Artificial Intelligence More Explainable}

%%
%% The "author" command and its associated commands are used to define
%% the authors and their affiliations.
%% Of note is the shared affiliation of the first two authors, and the
%% "authornote" and "authornotemark" commands
%% used to denote shared contribution to the research.
\author{Sopam Dasgupta}
\email{sxd180105@utdallas.edu}

\affiliation{%
  \institution{The University of Texas at Dallas}
  \streetaddress{800 W Campbell Rd}
  \city{Richardson}
  \state{Texas}
  \country{USA}
  \postcode{75080}
}

%%
%% By default, the full list of authors will be used in the page
%% headers. Often, this list is too long, and will overlap
%% other information printed in the page headers. This command allows
%% the author to define a more concise list
%% of authors' names for this purpose.
\renewcommand{\shortauthors}{Sopam Dasgupta}

%%
%% The abstract is a short summary of the work to be presented in the
%% article.
\begin{abstract}
Machines are being increasingly used in decision-making processes, resulting in the realization that decisions need explanations. Unfortunately, an increasing number of these deployed models are of a `black-box' nature where the reasoning behind the decisions is unknown. Hence there is a need for clarity behind the reasoning of these decisions. As humans, we would want these decisions to be presented to us in an explainable manner. However, explanations alone are insufficient. They do not necessarily tell us how to achieve an outcome but merely tell us what achieves the given outcome. For this reason, my research focuses on explainability/interpretability and how it extends to counterfactual thinking.
\end{abstract}

%%
%% This command processes the author and affiliation and title
%% information and builds the first part of the formatted document.
\maketitle

\section{Introduction}

In recent times, due to the strides made in \emph{machine learning} (ML) and \emph{artificial intelligence} (AI), the performance of such systems in real-world applications has seen exponential growth. As a result, many tasks that one would traditionally attribute to being done by a human being are being performed by machine learning (ML) and artificial intelligence (AI) based models. Hence it is not surprising to see machine learning models being deployed in areas where historically, due to the nature of the tasks, it would require the involvement of a human, e.g., getting a loan/ receiving a bail judgment. 

Unfortunately, many of these state-of-the-art machine learning or artificial intelligence-based systems are so complex that we are unable to understand \emph{why} they made such a decision. This lack of clarity contributes to such models being viewed as a black box whose content/logic is unknown.

A natural consequence of the increasing involvement of machine learning models in decision-making processes is that  these decisions either directly or indirectly impact individuals. If another person's decisions impacted us, we would want a justification for the decision, e.g., on receiving judgment from a judge. We, as humans, can accept decisions if they are justified. Therefore if an algorithm or model's decision cannot be interpreted, it cannot be trusted. 

 Due to the ever-increasing complexity of models and the \emph{complexity-explainability} trade-off, there has been a greater focus on making machine learning models more comprehensible. Hence, extending interpretability to machine learning models became a necessity. 

This survey explores the various approaches taken to solve this problem.

\section{Definitions}

As models are becoming increasingly more complex, allowing them to solve more challenging tasks, we have encountered a new problem- many of these complex models need to be explained to us as humans. There have been various approaches to tackling this problem and understanding the reasoning employed by these complex models. Some standard terms being used in this regard are \emph{explainability} and \emph{interpretability}, both of which have no concrete mathematical definition.

\emph{Interpretability} is defined as the degree to which humans can identify cause-effect relationships. As \emph{interpretability} increases, so does the ability to identify cause-effect relationships. On the other hand, \emph{explainability} is concerned with conveying the internal logic of the decision-making model. As \emph{explainability} increases, so does our understanding of the model's logic. For the purposes of the survey, we shall use them interchangeably.

We will categorize the various approaches to achieving \emph{explainability} into four types. The first is the \emph{intrinsic} and \emph{post hoc} explanation models. They are identified by how they achieve explainability. \emph{Intrinsic} models are self-explainable. They focus on explainability over accuracy., e.g., linear models like linear regression or tree-based models like decision trees. On the other hand, \emph{post hoc} explanation models focus on explaining already existing black box models. The methods for explaining black box models make use of a secondary \emph{intrinsic} model, e.g., LIME uses simple \emph{intrinsic} explanation models like linear models or decision trees to explain any underlying classifier. 

Approaches to explainability can also be categorized based on the flexibility of the approach. For example, \emph{model-agnostic} methods are not limited to any particular model type and can explain any model, e.g., LIME, CLEAR. On the other hand, as the name suggests, \emph{model-specific} approaches are limited to specific model types, e.g., Linear SHAP and DeepLIFT.

Models are also categorized based on the type of data they explain. Some models may be restricted to text-based or tabular data, while others to images. LIME  is a popular algorithm that can be used to explain tabular, textual, and image data. DeepLIFT can be used to explain image data.

The scope is essential, whether local or global. Models can be \emph{locally explainable} or \emph{globally explainable}. Local explainability refers to sufficiently explaining the predictions in the local region around the instance. It, however, cannot explain the global behavior of the model. \emph{Globally explainable} refers to explaining the working of the model as a whole, e.g., FOLD, FOLD-R++. The LIME  approach makes use of \emph{local explainability} to obtain \emph{global explainability}. 

The conventional term of accuracy is not enough for evaluating models that provide explanations. \emph{Fidelity} is defined as the measure of how well the explanations approximate the black box model. If accuracy and \emph{fidelity} are both high, then the explanations closely map the model. Conversely, if the \emph{fidelity} is low, then the explanations are useless. It will be shown later that one of the shortcomings of the popular LIME  algorithm lies in its poor \emph{fidelity}. 

One of the main reasons for research in explainable artificial intelligence (XAI) is to increase trust in the complex machine learning models being deployed. However, first, we must define what we mean by trust. Ribeiro et al. defined different notions of trust: \emph{trust in a prediction} and \emph{trust in a model}. While related, they are not the same. The result of their work,  LIME, uses trust in decisions to leverage trusting a model. 

Lundberg and Lee \cite{SHAP} introduce a new perspective of viewing any explanation of a model's prediction as a model itself. They defined it as an \emph{explanation model}. An explanation model is thus any interpretable approximation of the original model. They use simplified inputs that map to the original inputs through a mapping function. These simplified inputs are interpretable to humans. Lundberg and Lee \cite{SHAP} presented a unifying framework combining six explanation based methods that guaranteed three properties:
\begin{enumerate}
    \item \emph{Local accuracy} of the explanation model: Accuracy of the explanation model in modeling the underlying classifier around local regions
    \item \emph{Missingness}: Missing features should not effect the decision process.
    \item \emph{Consistency}: If the \emph{simplified input's} contribution increases or is unchanged, the inputs attribution should not decrease. \newline
\end{enumerate}Another term that will be encountered in this survey is that of \emph{counterfactuals}. It was proposed by Watcher et al. \cite{Watcher} as an alternate explanation without exposing the inner logic of the model. It is of the form: \emph{ Score p was returned because variables V had values (v$_1$, v$_2$, ...) associated with them. If V instead had values (v$_1^{\prime}$, v$_2^{\prime}$, ...), and all the other variables had remained constant, score p$^{\prime}$ would have been returned}. It imagines a world where the fact is not true, but the counterfactual is true. According to Watcher \cite{Watcher}, the earlier definitions of an \emph{explanation} referred to conveying the \emph{internal state or logic} of the algorithm. Counterfactuals, on the other hand, provide explanations in the form of a response to an external change. 

A related term to \emph{counterfactuals} is that of \emph{algorithmic recourse}. It is defined as the ability of a person to obtain a desired outcome. Its utility was highlighted by Ustun et al., who noted that some \emph{counterfactual} explanations are not realizable for specific individuals. e.g., the bank, on rejecting the loan application, justifies it by saying- if you reduce your age, you will receive the loan. This problem arises as counterfactuals do not take into account the world we live in and the rules governing it. They specifically highlighted how some counterfactuals were not realizable. They categorized the features into various types and highlighted how recourse depends on the algorithm's features
\begin{enumerate}
        \item \emph{Actionable}: These feature can be directly acted or intervened upon. e.g., An increase in salary will result in the bank loan being approved.
        \item \emph{Immutable}: These feature cannot be altered. e.g., a change in the ethnicity  will result in the bank loan being approved.
        \item \emph{Conditionally mutable}: These features cannot be acted upon \emph{directly} but \emph{indirectly}. e.g., an increase in the credit score will result in the bank loan being approved. Credit score can be increased by never missing a payment of dues. 
\end{enumerate} 
One of the ways in which we can indirectly affect some features is by making use of the causal dependencies governing the world in which they lie. The work by Pearl \cite{Pearl} introduced a method to express the world through structural equations that model the causal dependencies in the world. These so-called \emph{Structural Causal Models} (SCM) represent the world and any change in the form of a counterfactual is the cause of an \emph{intervention} that models how the world will change depending on the altered feature.

Karimi et al. \cite{MACE, MACE_R} improved on the work by Ustun et al. by highlighting the use of \emph{counterfactual space}, which is defined as the space governing all possible inputs for a given model that would cause it to give a different outcome. However, they also highlighted how just the \emph{counterfactual space} is not enough to suggest realizable counterfactuals, e.g., a decrease in age will result in the bank loan being approved. Hence they introduced \emph{plausibility}, \emph{feasibility}, and \emph{diversity} constraints to generate appropriate counterfactuals and \emph{flipsets}.

\begin{figure}[htp]
    \centering
    \includegraphics[width=11cm]{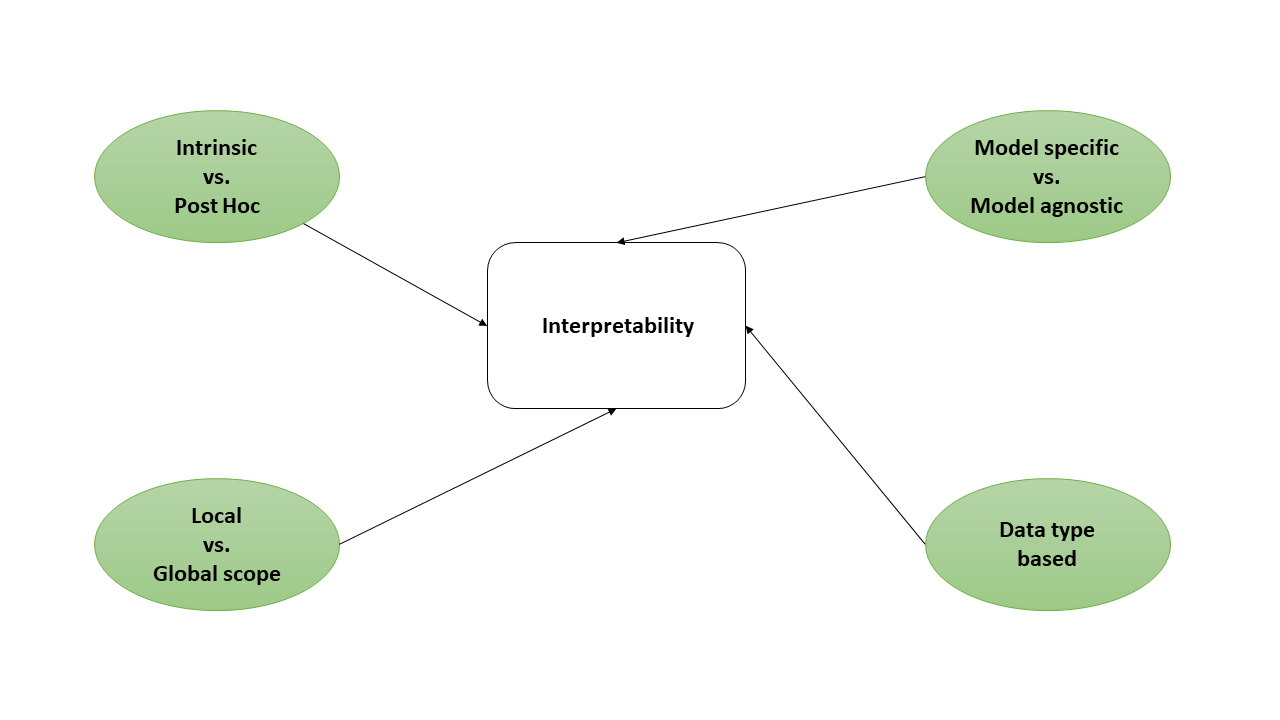}
    \caption{Types of Interpretable Models}
    \label{fig:Types_XAI_1}
\end{figure}

\section{Motivation}

Machines are being increasingly used for decision-making purposes. One of the natural effects of this is the fact that human lives are increasingly being affected by algorithmic decisions, either directly or indirectly. Some notable instances include algorithms for judging the value of a family home which, when deployed on a mass scale, have affected the rate of home ownership in several towns. Another example lies in the task of approving a loan in a bank. The bank makes use of algorithmic decisions to decide whether to approve or reject a loan application. 

For a long time, machine-learning models could not compete with humans in various tasks, e.g., recognizing objects in images, identifying real images from fake ones, and identifying diseases. With the advent of deep learning, there has been extensive development of a whole category of complex machine learning models. These models have started to outperform humans as well. However, a natural consequence of these complex models is that they are not easy to understand and hence not \emph{interpretable}. This is not just limited to deep learning based approaches. This lack of understanding/interpretability affects the trust that we, as humans, can place in the decisions. Hence there was a need to not just rely on the model's performance but also \emph{trust} the model and its decisions. 

\subsection{Different Approaches to Explainability}

The focus has been on providing explanations to achieve trust for the decisions being made. However, if we want decisions to be explainable, we need to define what is a good explanation. Bringing up what is considered to be a good explanation: \emph{it should answer why it predicted a certain output}, and \emph{it should inform on what changes should be made to obtain the desired output}. This was addressed in the work of Ribeiro et al. \cite{LIME}. They noted different notions of trust, namely \emph{trusting a decision} and \emph{trusting a model}. 

\subsubsection{Intrinsic  and post hoc explanation models} The novel algorithm LIME, resulting from the work of Ribeiro et al. \cite{LIME}, is able to provide explanations for not only a decision but also for the whole model. It does so by learning \emph{local explanations} for individual decisions and leveraging the \emph{local explanations} to explain the model as a whole. LIME comes under a category of explainable models known as \emph{post hoc explanation}.
\emph{Post hoc explanation models} focus on explaining already existing models that are too complicated for human understanding. These underlying models are referred to as \emph{black box models}. The methods for explaining black box models make use of a secondary model/technique, which comes under another category of explainable methods known as \emph{intrinsic models}. 

\emph{Intrinsic explanation models} focus on the model's explainability over its performance. A simple model is preferred over a complicated model created to provide explanations. Some examples are linear models like linear regression and tree-based models like decision trees are also \emph{intrinsic} explanation models. They also include graph-based models like Bayesian networks in prediction tasks, e.g., Naive Bayes is \emph{intrinsic} as it calculates the probability of each feature independently, and this independence assumption highlights the contribution of each feature toward the decision. 

\subsubsection{Local and Global Explanation Models} 
We have spoken about LIME \cite{LIME} and its advantages. However, Lundberg and Lee \cite{SHAP} highlighted some of the shortcomings of LIME. Namely, it highlighted how LIME, as described by Ribeiro et al. \cite{LIME}, either violated the property of \emph{local accuracy} or the property of \emph{consistency}. Lundberg and Lee \cite{SHAP} introduced a framework to unify the approaches that come under the class of additive feature attribution methods which includes LIME. \emph{SHapley Additive Model Agnostic Explanations(SHAP)} \cite{SHAP} produces explanations with the following properties of \emph{local accuracy}, \emph{missingness}, and \emph{consistency}. Since LIME, while being an addictive feature attribution method, did not make use of the unique solution in the form of \emph{SHAPLEY values}, it violated either one of \emph{local accuracy} or \emph{consistency}. However, Lundberg and Lee \cite{SHAP} did extend LIME to satisfy the requirements in the form of \emph{KernelSHAP}. It was even highlighted that explanations possessing all 3 properties were more in line with human explanations compared to the model in the work of LIME by Ribeiro et al. \cite{LIME}. Explanations are only useful if they are correct. LIME \cite{LIME} fails to produce local explanations that agree with the model being explained. In other words, the \emph{local fidelity} was inadequate. This can be explained by the findings of Lundberg and Lee \cite{SHAP}, which say that LIME \cite{LIME} should violate either property of \emph{local accuracy} or \emph{consistency}. 

LIME produces \emph{local explanations} to achieve \emph{global explainability}.  Tree Explainer is a model with \emph{global explainability}. It is a \emph{model-specific} extension of SHapley Additive Model Agnostic Explanations \cite{SHAP_ref_1} for trees which was introduced by Lundberg et al.\cite{SHAP_ref_1}. It tries to obtain a global understanding through local explanations. Some other examples are \emph{First Order Learning Defaults (FOLD)} family of algorithms \cite{Fold, Foldrpp}. \emph{Counterfactuals} are another example of explanation as depicted by Watcher et al \cite{Watcher}. They are instance specific and hence \emph{local}. They provide explanations in the form of the minimal change necessary to achieve an alternate outcome. 

\subsubsection{Model-Specific and Model-Agnostic Explanation models}
LIME was extended with \emph{Shapley values} to produce KernelSHAP which is a \emph{model-specific explanation method}. Unlike \emph{model-agnostic methods}, these are usually constrained to certain models of families of models, e.g., Tree Explainer, a \emph{model-specific} extension of SHAP for trees was introduced by Lundberg to understand global model structure using local explanations. LIME, SHAP, CLEAR \cite{CLEAR} are all examples of \emph{model agnostic explanation models}. As their name suggests, they are \emph{model-agnostic} and can hence be applied to explain any model. 

Shakerin \cite{Fold}, Wang and Gupta \cite{Foldrpp} took a different approach to solving the problem of understanding the logic of models. Due to the lack of intuitive descriptions making justifying a decision harder, they developed rule-based approaches to achieve explainability. They make use of Inductive Logic Programming (ILP) using Horn clauses that are comprehensible to humans. It has the advantage of being flexible with regard to adding more background knowledge without relearning the entire model. They showed how to achieve explainability in the form of default rules. Default rules which follow common sense reasoning are easily understandable to humans. Their algorithms FOLD \cite{Fold}, and FOLD-R++ \cite{Foldrpp} are some examples of \emph{model-agnostic} rule based explanation methods.

\subsection{Provide Explanations Through Suggestions- Counterfactuals}

The notion of counterfactuals as explanations for automated decisions was proposed by Watcher et al. \cite{Watcher}. Earlier, explanations were thought to provide an understanding of the internal logic of the model. Counterfactual explanations on the other hand describe a dependency on the external factors that led to the decision. This was used as an alternate solution to the problem of explaining the model. Counterfactuals provide explanations that are meaningful for individual decisions with regards to 1) understanding why the decision was made, 2) how to contest the judgment/decision and finally 3) advice on how what to do to obtain the desired judgment/decision. Watcher et al. \cite{Watcher} highlighted the utility of counterfactuals as explanations by satisfying the above requirements. 

To address the problem of local accuracy in LIME \cite{LIME}, White et al. \cite{CLEAR} made use of \emph{counterfactual explanations}. They are of the form - \emph{ Score p was returned because variables V had values (v1, v2, ...) associated with them. If V instead had values (v1$'$, v2$'$, ...), and all the other variables had remained constant, score p$'$ would have been returned}. They are used to provide instance based explanations and can also be used to model causal relations.

White et al.\cite{CLEAR} combined the features of an explanation model like LIME  with those of \emph{counterfactual explanations} to produce a novel method. Since \emph{Shapley values} are applied to binary features, they cannot be used to estimate the effects of changing a numerical feature. Hence they cannot be used to estimate counterfactuals. Extending LIME \cite{LIME} by making use of counterfactuals as depicted by Watcher et al. \cite{Watcher}, White et al. \cite{CLEAR} improved the \emph{local accuracy} of LIME to create CLEAR \cite{CLEAR}. CLEAR \cite{CLEAR} made use of counterfactuals to improve the local model for explanation. As a result it has improved \emph{fidelity} over LIME \cite{LIME}. In other words, the \emph{intrinsic explanation model} used closely approximates the underlying black-box model. Therefore this highlights its advantage over the popular LIME \cite{LIME} approach by utilizing \emph{counterfactual explanations}. Similar to LIME and SHAP, CLEAR \cite{CLEAR} is also \emph{model agnostic}. 

Byrne \cite{Byrne19} took a different approach. Exploring how counterfactuals relate to the human thought process, they linked counterfactual explanations to how humans think. Byrne \cite{Byrne19} even highlighted how humans create counterfactuals and the common fault lines in doing so. This knowledge does provide some direction in knowing how we as people go about generating counterfactuals which can in turn be used to help with the explanation problem, e.g., If your credit score was above 750, the loan would have been approved. However Watcher et al. \cite{Watcher} noted that the counterfactuals created didn't take into account causal dependencies governing the world.

What would happen if the counterfactuals created suggested changes that are impossible to realize? Ustun et al. \cite{Rec} highlighted the problems caused by such impossible suggestions, e.g., You would have obtained the loan if you were a Male. Impossible suggestions are of no utility. Ustun et al. \cite{Rec} focused on generating classifiers that achieve recourse for all individuals. They provided tools and came up with a method to generate counterfactuals explanations that are possible for the individual to realize. Ustun et al. \cite{Rec} highlight the feasibility and difficulty of recourse of some counterfactuals. In addition it showed how the \emph{mutability} and \emph{actionability} of features have an effect on \emph{algorithmic recourse}. Their work also helps evaluate the quality of counterfactuals. In addition, the work by Ustun et al. \cite{Rec} also provides alternate counterfactual solutions and possible ways to achieve the desired outcome, referred to as \emph{flipsets}. These \emph{flipsets} act as suggestions on how to achieve the desired result. Unfortunately, the scope of their work was limited to linear classification problems.

The work by Karimi et al. \cite{MACE} provided a \emph{model-agnostic} approach in the form of logical formulae. It achieved the task of finding the nearest counterfactuals explanations by taking lessons on mutability and actionability of features from Ustun et al. \cite{Rec}.

However the earlier works on recourse did not take into account the nature of the world we live in. So the notion that the \emph{nearest counterfactual explanation} will require the minimum effort from the subject is \emph{not true}. This is due to the fact that the approaches did not taken into account the causal relationships that govern the world. This was corrected in the work by  Karimi et al. \cite{MACE_R} who highlighted the vulnerability of \emph{recourse} to such causal relations. They provided a novel approach in the form of interventions \cite{Pearl}. This new approach of counterfactuals through interventions has the advantage of modeling the changes caused by the causal dependence of the features.

In this survey we shall cover the works done on \emph{explainability} in \emph{artificial intelligence} and \emph{machine learning}. In addition to this, we shall explore some of the popular algorithms for providing explanations and where the direction of research is heading.

\section{Explainability in AI}

The involvement of machines in a whole host of avenues highlights their increasing involvement in our lives. One of the noticeable changes lies in how roles or services which we assumed could only be performed by humans are now being performed in a manner just as good if not better than humans, e.g., a bank manager approves a loan application vs. an algorithm approves a loan application. A benefit of this would be the improved performance in tasks. This benefit also comes with its consequences, as we shall explore.

\subsection{Using Explainability as a Means to Develop Trust}
 If decisions being made affect us negatively, that will be a problem. When people make decisions, they are required to justify them. This promotes a sense of trust. However, now the decisions are being made by machines/algorithms. Hence the task of justifying the decisions becomes more complicated, especially when the decision-making model is far too complex for human understanding. This causes an issue- \emph{how can we trust such a model} and, by extension, the decisions made by it? 
 
 Ribeiro et al. \cite{LIME} try to achieve this absence of trust through \emph{explanations}. For trust in a decision and, by extension, the model making the decisions, one should be able to explain the decisions being made. This brings up the question of what is considered to be a good explanation 1) it should answer why it predicted a certain output, and 2) it should inform on what changes should be made to obtain the desired output.
 
 This is slightly different from trust. Ribeiro et al. \cite{LIME} highlighted different notions of trust 1) \emph{trust in a decision made} and 2) \emph{ trust in the model that makes decisions}. Ribeiro et al. \cite{LIME} takes the approach of explaining individual predictions as well as multiple predictions as a solution to \emph{trusting a prediction} problem and the \emph{trusting a model} problem. 
 
 Some of the notable contributions of their work was the creation of \emph{Locally Interpretable Model-Agnostic Explanations (LIME)}. It explains the prediction of \emph{any} model. It comes under the category of a \emph{post hoc explanation model}. It makes use of an \emph{intrinsically explainable model} such as linear models, decision trees or falling rule lists \cite{LIME_ref_3} to explain an individual prediction locally. As the name suggests, it is \emph{model-agnostic} and can even be used to explain various data types such as visual and textual artifacts and their relation to predictions. 
 
 To understand the global behaviour of the underlying model, SP-LIME was introduced. It explains a set of representative instances as a solution to provide \emph{global explainability}. By using diverse, representative instances, SP-LIME explains how the model behaves globally and hence helps in the \emph{trusting a model} problem. By learning local explanations for individual decisions and leveraging the local explanations to explain the model as a whole, LIME \cite{LIME} is able to provide global explainability. As a consequence of using an \emph{intrinsically explainable model}, if the underlying model too complex in the locality of the prediction, then the explanations produced by the \emph{intrinsic model} may not be faithful, e.g., if the intrinsic model is linear and the underlying model is non linear locally, the explanations will have little utility. While this is the limitation of LIME, its has the advantage of being \emph{model agnostic}.
 
 LIME allows the evaluation of decisions and models based on trust. This is especially helpful in cases where one needs to choose a model from a diverse set having similar accuracy. This evaluation is done based on trustworthiness. We can leverage the explanations provided by LIME to enhance the \emph{trust in the model} and help decide amongst them. Explanations from LIME were found to have helped non-experts pick which explanation model generalises better on unseen data.
 
 LIME was found to provide the added benefit of improving untrustworthy classifiers. By being given explanations, even non-experts can modify the learning process (removing untrustworthy features from learning) to create trustworthy classifiers. 
 
 It outperforms methods like parzen \cite{LIME_ref_1}. Even against a greedy procedure \cite{LIME_ref_2}, it is far more trustworthy.

\begin{figure}[htp]
    \centering
    \includegraphics[width=07cm]{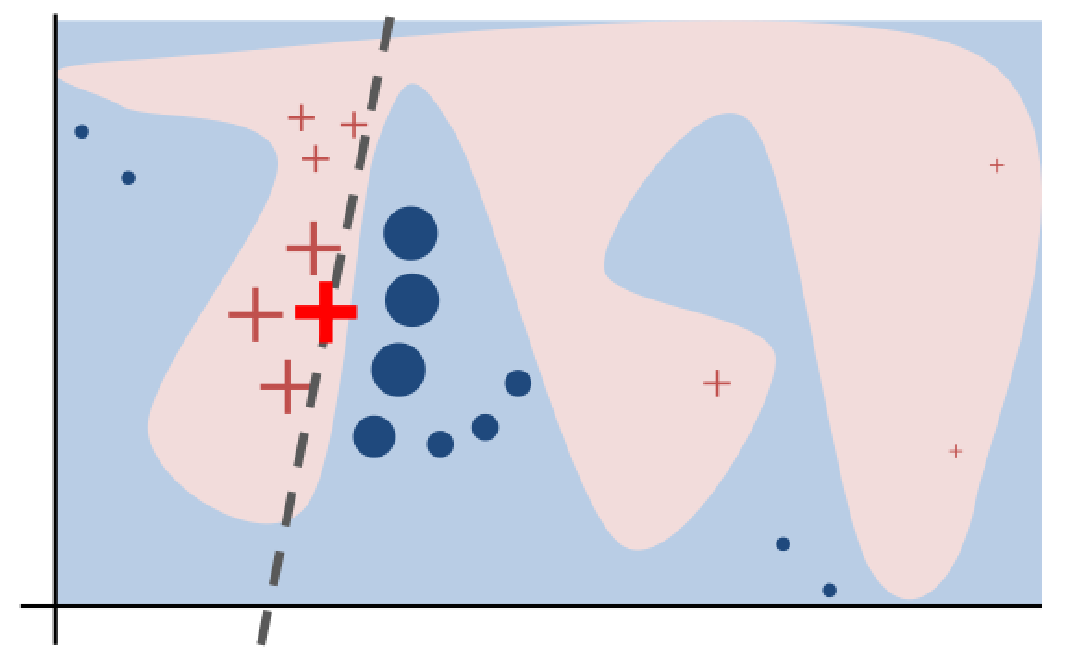}
    \caption{The dotted line is the linear classifier for explaining the instance}
    \label{fig:Types_XAI_2}
\end{figure}

\begin{table}[htbp] 
\begin{tabular}{||c c c c c c||} 
 \hline
  & dataset & LR & NN & RF & SVM\\ [0.5ex] 
 \hline\hline
 Random & Books & 14.6 & 14.8 & 14.7 & 14.7\\ 
 \hline
 Parz & Books & 84.0 & 87.6 & 94.3 & 92.3\\ 
 \hline
 Greedy & Books & 53.7 & 47.4 & 45.0 & 53.3\\ 
 \hline
 LIME & Books & 96.6 & 94.5 &96.2 & 96.7\\
 \hline
 Random & DVDs & 14.2 & 14.3 & 14.5 & 14.4\\ 
 \hline
 Parzen & DVDs & 87.0 & 81.7 & 94.2 & 87.3\\ 
 \hline
 Greedy & DVDs & 52.4 & 58.1 & 46.6 & 55.1\\ 
 \hline
 LIME & DVDs & 96.6 & 91.8 & 96.1 & 95.6\\[1ex] 
 \hline
\end{tabular}
\caption{\label{Table_LIME} F1 score depicting trust for different explanation methods on a collection of classifiers (Logistic Regression, Neural Network, Random Forest and SVM) and datasets}
\end{table}

\subsection{Shortcomings of Explanations}

Through Ribeiro et al. \cite{LIME}, we have a \emph{model-agnostic} method for obtaining explanations. However, explanations are only useful if they are correct. Explanations that do not agree with the model have no use. This is highlighted through the work of White et al. \cite{CLEAR}. It introduced the notion of \emph{fidelity to the underlying classifier} and highlighted LIME's limitation \cite{LIME} of producing unfaithful local explanations. This was indicated through an earlier work by Lundberg and Lee \cite{SHAP}. It introduced a unified approach for interpreting model predictions, SHAP (SHapley Additive exPlanations). It subsumes approaches such as \emph{LIME} \cite{LIME}, \emph{DeepLIFT} \cite{shap_org_7, shap_org_8} and \emph{layer-wise relevance propagation} \cite{shap_org_1}. Additionally, it unifies approaches that make use of equations from game theory to provide explanations, .eg., \emph{Shapley regression values} \cite{shap_org_4}, \emph{Shapley sampling values} \cite{shap_org_9} and \emph{Quantitative Input Influence} \cite{shap_org_3}.  The resulting unified approach SHAP (SHapley Additive exPlanations) guaranteed the properties \emph{local accuracy}, \emph{missingness}, and \emph{consistency}. As long as a model is an \emph{additive feature attribution} method, they can be extended with this approach. Linear LIME \cite{LIME} which uses an linear model as the \emph{explanation model} is an \emph{additive feature attribution} method. 

The work by Lundberg and Lee \cite{SHAP} highlighted that the only possible setting in which all three properties would be obeyed by the model would be through SHAP values. It highlighted how the LIME algorithm by Ribeiro et al. \cite{LIME} either violates the property of \emph{local accuracy} and/or \emph{consistency}. The SHAP values are a set of values that provide a unique solution satisfying all three properties. The work of Lundberg and Lee \cite{SHAP} incorporated them into Linear LIME \cite{LIME}, to create Kernel SHAP. Kernel SHAP takes advantage of the three properties while still remaining \emph{model agnostic}. 

They also developed faster model specific approaches like Linear SHAP, Low-Order SHAP, Max SHAP, Deep SHAP. Like how Kernel SHAP incorporates Shapley values into LIME to satisfy the properties of \emph{local accuracy}, \emph{missingness}, and \emph{consistency}, Deep SHAP does the same but for the DeepLIFT algorithm. DeepLIFT\cite{SHAP_ref_2} approximates SHAP values. Since DeepLIFT is an \emph{additive feature attribution method} that satisfies \emph{local accuracy} and \emph{missingness}, \emph{Shapley values} represent the unique solution to satisfy consistency. DeepSHAP is hence an extension of SHAP values to DeepLIFT.

On comparing Kernel SHAP, Shapley sampling values and LIME,  we see that Kernel SHAP identifies more accurate feature importance estimates with fewer evaluations. This highlights the advantage of the guarantees of \emph{local accuracy, missingness, and consistency} that is present in the Kernel SHAP.

 We have a goal of producing human interpretable explanations. We need to ask if these properties also guarantee better, more human-intuitive explanations. In work by Lundberg and Lee \cite{SHAP}, this was explored and found to be accurate. The explanations arising from models possessing all three properties were more in line with human explanations compared to LIME.The on comparing Kernel SHAP with DeepLIFT, we get the same results- Kernel SHAP produces more human-intuitive explanations.

Highlighting how some of the earlier work, including LIME, were all a part of a class of methods called \emph{additive feature attribution methods} which have an explanation model that is a linear function of binary variables. It shows how using \emph{Shapley values} in additive feature attribution methods achieve the given properties and hence produces better explanations.

\begin{figure}[htp]
    \centering
    \includegraphics[width=13cm]{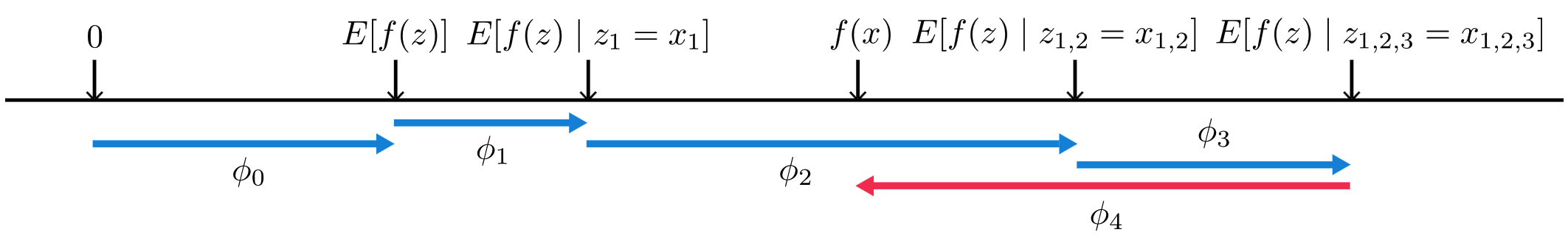}
    \caption{ SHAP (SHapley Additive exPlanation) values attribute to each feature the change in the
expected model prediction when conditioning on that feature}
    \label{fig:Types_XAI_3}
\end{figure}

\subsection{Explanations using Defaults and Exceptions}

Different people explored different approaches to tackle the task of explainability. The approach of Shakerin \cite{Fold} was to discover the logic of the underlying model in the form of a set of rules. There were some advantages to this approach. Namely, the rules are highly intuitive to understand. This understanding of the explanation rules is easy to follow and could lead to further action, e.g., this allows a bank to inform an applicant what they need to obtain the desired outcome. The reason why these rules are so intuitive is because of the use of \emph{defaults} and \emph{exceptions}. Earlier work in which FOLD is an improvement on such as FOIL \cite{FOIL}, produces rules that could be more intuitive to us. 

\begin{figure}[htp]
    \centering
    \includegraphics[width=9cm]{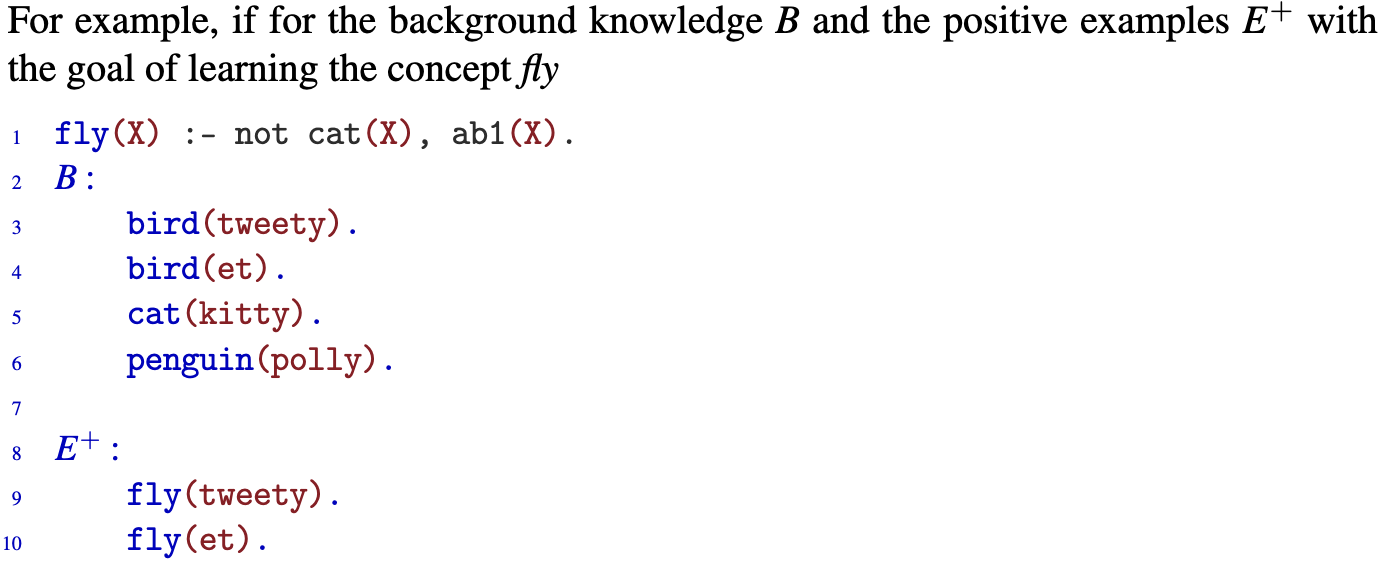}
    \label{fig:KB}
\end{figure}

\begin{figure}[htp]
    \centering
    \includegraphics[width=9cm]{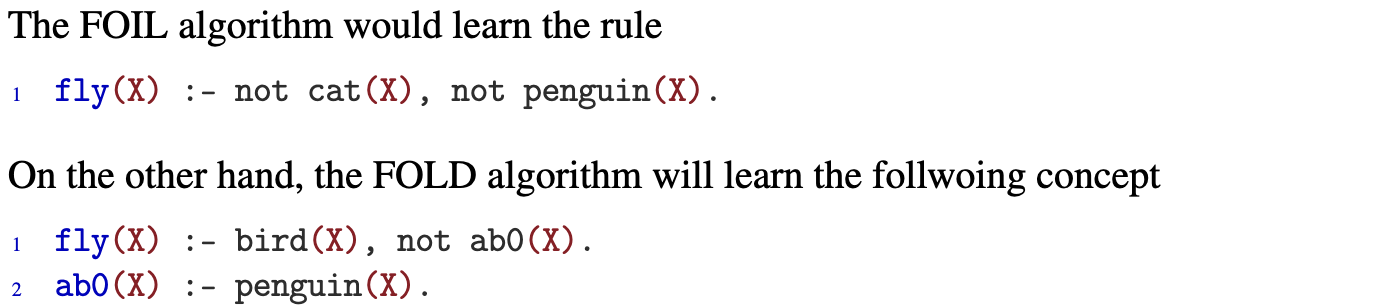}
    \label{fig:FOLD_FOIL_RULES}
\end{figure}

As humans, we tend to think in the form of default rules and exceptions to those defaults.
This form of explanation is more intuitive than the one in FOIL.

The work by Shakerin \cite{Fold} improved on the shortcomings of Inductive Logic Programming (ILP), which is unable to handle exceptions to rules. Being unable to handle exceptions is a problem, as that is not how humans think. Inductive Logic Programming (ILP) was incorporated with \emph{negation-as-failure} \cite{FOLD_ref_2}. The resultant theory - \emph{default theory} resembles the common sense reasoning employed by humans, thereby being more intuitive. 

\begin{figure}[htp]
    \centering
    \includegraphics[width=11cm]{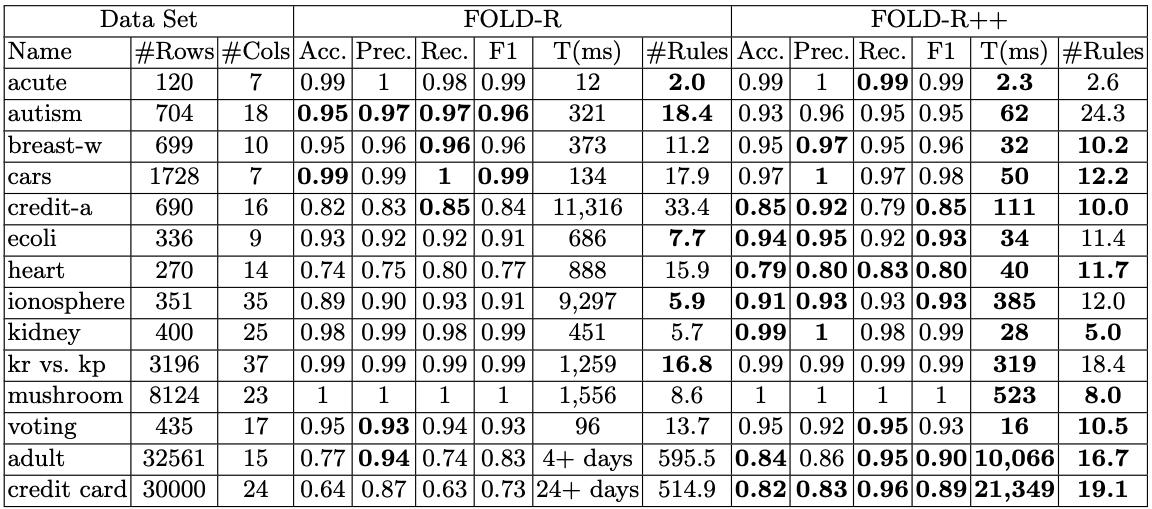}
    \caption{Comparison of FOLD-R and FOLD-R++ on various Datasets}
    \label{fig:FOLD_vs_Rpp}
\end{figure}
Another advantage or feature of thinking in the form of \emph{defaults} and \emph{exceptions} is that it allows reasoning in the absence of information, e.g., if John's friend is unable to meet him for tea, John reasons that they are either visiting the hospital or might even be asleep. 

The similarities to how humans think do not end there. Many traditional machine learning (ML) approaches suffer from the problem of having to relearn the whole model in the presence of new information. This is a problem as it can be extremely time-consuming, and it does not reflect human thinking. As people, we do not have to relearn everything on encountering new knowledge. One of the advantages of this approach is that it allows background knowledge to be incrementally extended without relearning the whole model. This is identical to how we think.

Shakerin proposed the First Order Learner of Default-theories algorithm (FOLD) \cite{Fold}. It is an improvement on an existing popular algorithm called FOIL \cite{FOIL}. FOLD learns default theories from background knowledge and positive and negative examples. It learns in the form of \emph{defaults} and \emph{exceptions}. It goes about it by first specializing in a general rule with positive literals. Each hypothesis learned, unless perfect, will learn some negative examples. Each specialization should rule out some negative examples without reducing the positive examples covered significantly. Exceptions could be learned by swapping current positive and negative literals and learning the exceptions. This recursive approach allows FOLD to learn exceptions to exceptions.

\begin{figure}[htp]
    \centering
    \includegraphics[width=11cm]{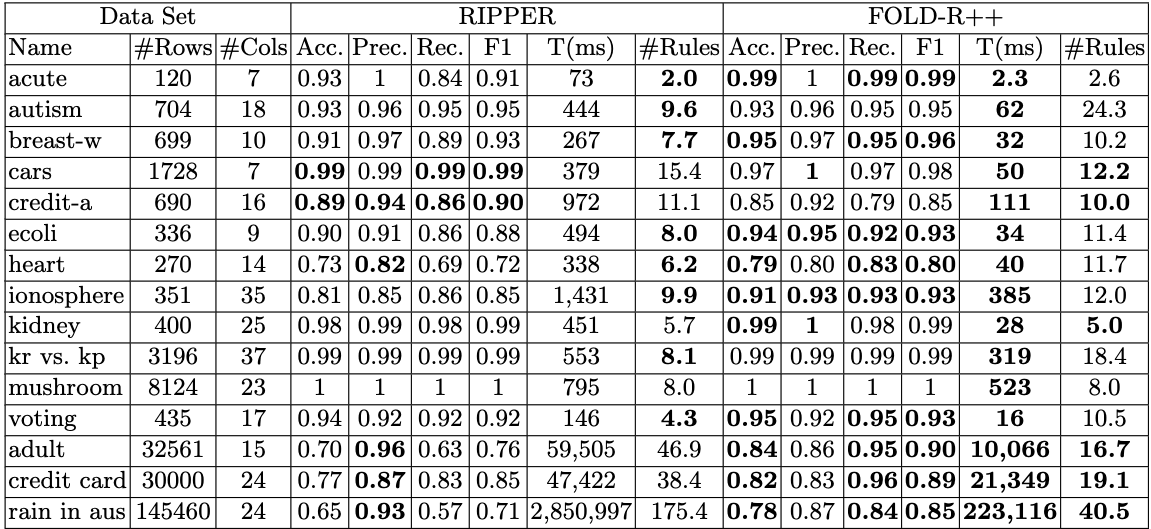}
    \caption{Comparison of FOLD-R++ and RIPPER on various Datasets}
    \label{fig:FOLDRpp_vs_RIPPER}
\end{figure}
Shakerin also expanded on LIME\cite{LIME} and SHAP\cite{SHAP} through LIME-FOLD\cite{LIME_FOLD} and SHAP-FOLD. By making use of LIME \cite{LIME} to compute the features responsible for decisions, this knowledge was incorporated into FOLD \cite{Fold}. The result of this was the production of more human-intuitive explanations in the form of a smaller number of rules. To ensure the benefit of the three properties of \emph{local accuracy}, \emph{missingness}, and \emph{consistency}, FOLD was incorporated with \emph{Shapley values}. The resulting algorithm groups similar data samples under the same clause. The resultant algorithm SHAP-FOLD could also be used to capture the global behavior of statistical models. 

FOLD, for all its advantages, suffered from some major drawbacks- namely, its inability to handle numeric features directly. Shakerin later extended the FOLD algorithm to alleviate this problem in the form of FOLD-R, sacrificing the efficiency and scalability of the FOLD algorithm. 

Wang and Gupta \cite{Foldrpp} recognized this problem and proposed a novel algorithm FOLD-R++. The core tool behind this algorithm was the proposed novel \emph{prefix-sum} technique. The \emph{prefix sum} technique targets the slowest, most time-consuming component of the FOLD family of algorithms that FOLD-R++ is a part of, namely the calculation of the information gain heuristic. Information Gain (IG) is a heuristic used by many algorithms (decision trees). By introducing the novel technique, not only can the heuristic calculation be significantly sped up, thereby improving the efficiency, but the hyper-parameter \emph{ratio} introduced through this technique allows flexibility in controlling the nesting of exceptions. In addition to this, FOLD-R++ does allow negated literals in the default portion of its rules. This is unlike FOLD and FOLD-R, where it cannot, regardless of the correctness of the rules. FOLD-R++'s advantages in terms of efficiency and scalability are especially highlighted in larger datasets.
\begin{figure}[htp]
    \centering
    \includegraphics[width=9cm]{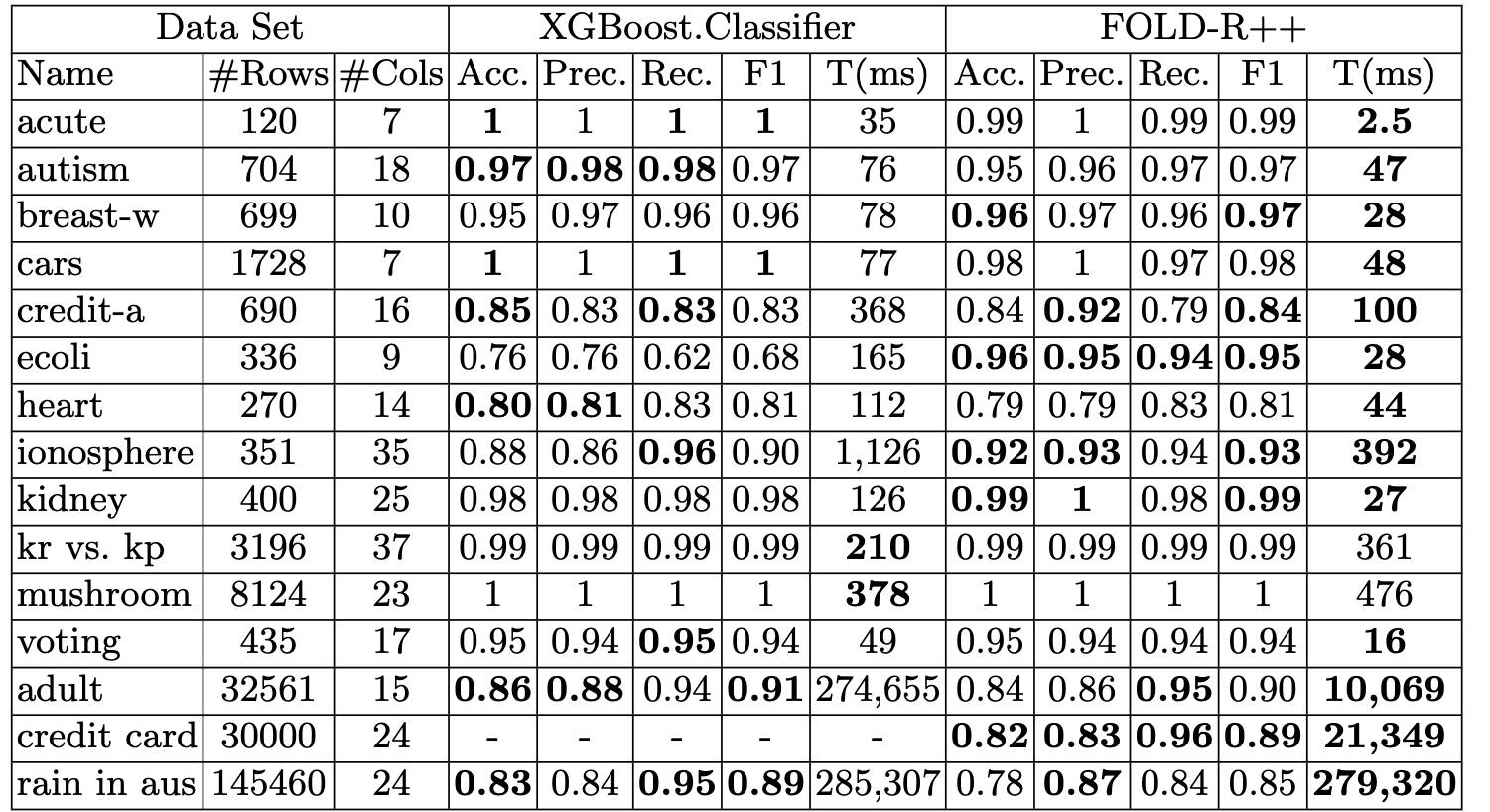}
    \caption{Comparison of FOLD-R++ and XGBoost on various Datasets}
    \label{fig:FOLD_vs_XG}
\end{figure}

\subsection{Consequences of Bad Explanations}

One of the goals of Explainable AI is to help people. In this sense, the focus should lie on how the explanations provided end up impacting us. Many of the works that have been proposed deal with explaining already existing models. However, Nourani et al. \cite{Gog} highlighted that the focus should instead lie in understanding how these explanations impact user behavior. They specifically highlighted how \emph{explanation veracity} affects user performance and agreement in machine learning and artificial intelligence models.

Nourani et al. \cite{Gog}, through a controlled user study, explored how human behavior is affected by decisions either with high, low veracity explanations or no explanation altogether. They even explore human behavior with no AI-suggested decision. Their experiments provided users with various queries. There were various cases 1) decisions accompanied with high veracity explanations that accurately explain what is going on, 2) decisions accompanied with low veracity explanations that inaccurately explain what is going on, 3) decisions and no explanation, and 4) no system decision, just the user's personal opinion. The aim was to see how the user's \emph{trust} in the system is impacted. Their hypothesis was that low \emph{veracity} explanations would contribute to greatly harming the trust in a model. 

Their work refers to the alignment of truthful and accurate representation of machines as \emph{veracity}. The intended goal is to study the understanding of a machine learning (ML) / artificial intelligence (AI) system and its effectiveness among a non-specialist population that is devoid of domain expertise or AI-related knowledge. Their proposed system outputs a human-understandable explanation using a two-layer architecture with an interpretable model on top of a deep uninterpretable layer.

Nourani et al. \cite{Gog} had three goals: 1) To check if their explainable AI model with explanations did indeed improve the user's performance, 2) To explore how the presence and veracity of explanations, whether good or bad affect the user's performance in decision making as well as their trust in the model. Finally, they intend to explore how the perceived accuracy and mental model of the intelligent systems are affected by these explanations.

The authors observed for time and error metrics the \emph{high veracity} explanations accompanying decisions resulted in significantly better performance compared to those without explanations or those with no AI decision provided. This helped answer the question of whether their system does indeed help improve user performance. The effect of the type of veracity was also significant. It was observed that \emph{low veracity} was significantly less accurate compared to those with not only \emph{high veracity} explanation accompanying decisions but also those without explanations provided or those with no AI-assisted decision given.

\begin{figure}[htp]
    \centering
    \includegraphics[width=11cm]{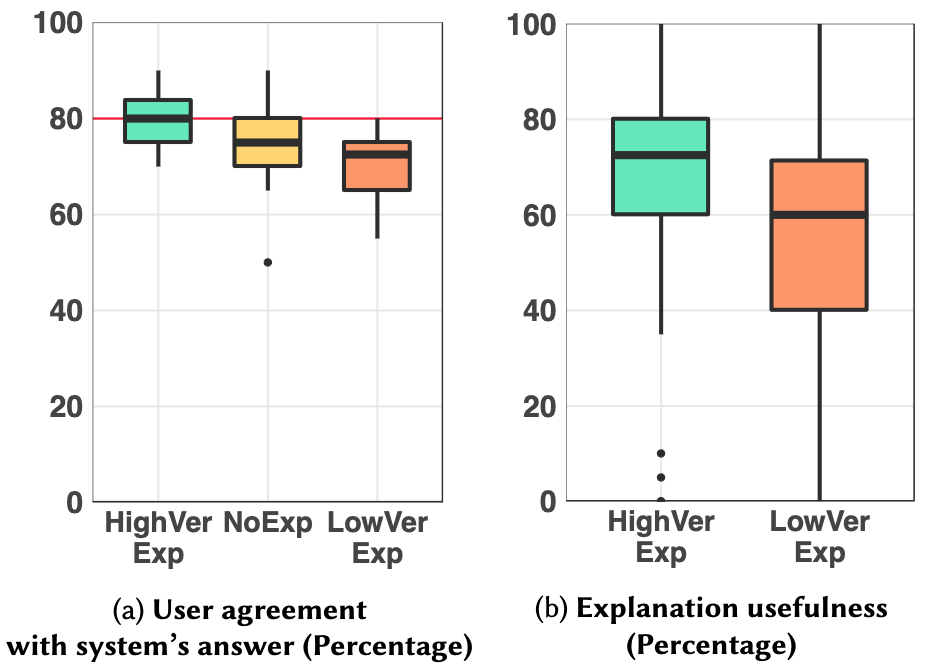}
    \caption{The left plot shows the percentage of times the answer of the user matched the system’s answer. The right plot shows the number of times users found explanations helpful in the explanation conditions. }
    \label{fig:Gog_org_2}
\end{figure}

The authors wanted to explore how the veracity of the explanations provided affected user agreement with the model. They saw that the highest level of user agreement with the model's decisions was found when decisions accompanied high \emph{veracity} explanations. The lowest level of user agreement was found when decisions accompanied low \emph{veracity} explanations. Their results showed that the low veracity explanations shower overall lower agreement with the system's decisions. It suggested that these explanations contributed to the loss of trust in the model regardless of it the model was indeed incorrect. This is significant as it highlights that user disagreements with the systems are related to the perceived accuracy of the system and are an indication of reduced trust in the system. 

The authors explored the user's mental model or user's understanding of the system and perception of accuracy in the XAI system. The task was to try and predict the system's decisions. This was done for high veracity, low veracity, and no explanation conditions. Surprisingly, the results did not indicate that explanation veracity and presence affected the user's ability to predict the AI model's decision output accurately. 

The work by Nourani et al. \cite{Gog} highlighted the impact of the quality of explanations provided on user behavior. Noting how \emph{veracity} of explanation did help in user agreement however the understanding of the model was not affected enough to provide a significant effect on the prediction of model output or accuracy. 

To test the effect of \emph{veracity}, the three conditions of \emph{high veracity}, \emph{low veracity}, and no explanations. Results indicated that user performance was significantly better for high-veracity explanations compared to low-veracity explanations. The lowered performance of low veracity explanations, even compared to no explanations, highlighted the power of explanation to influence the trust that users place in systems, regardless of the accuracy of the underlying system.
\begin{figure}[htp]
    \centering
    \includegraphics[width=13cm]{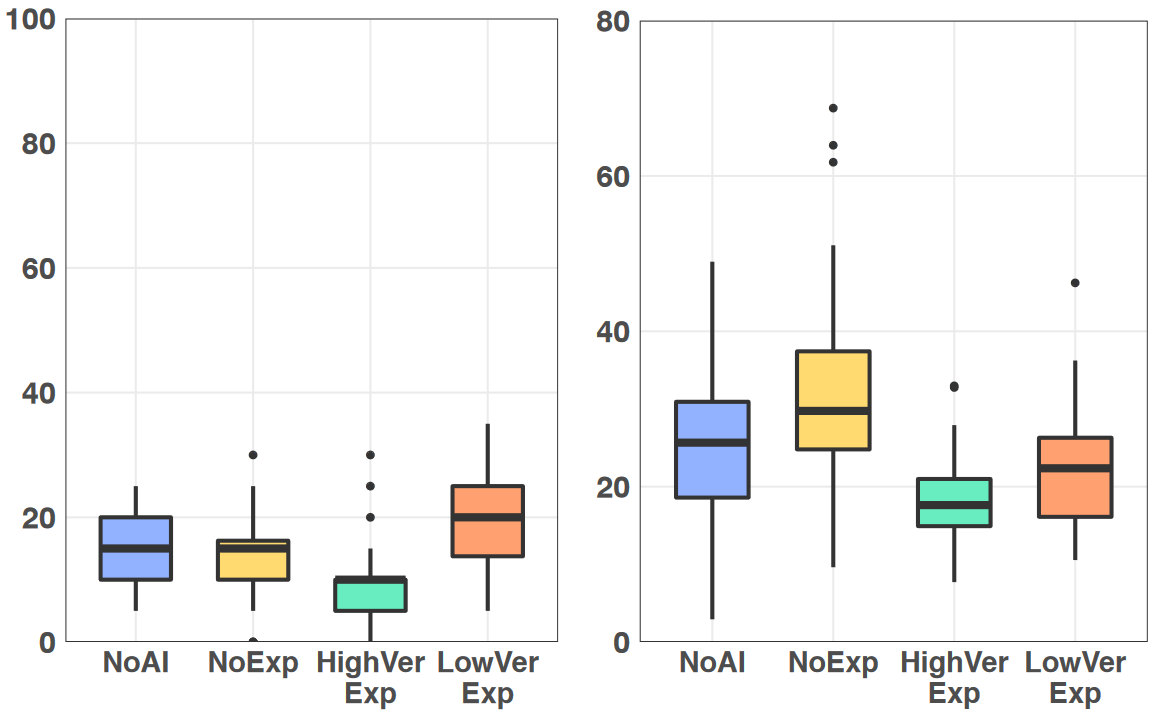}
    \caption{User performance for error on the left and time taken on the right.}
    \label{fig:Gogate_1}
\end{figure}

\subsection{Introduction to Counterfactuals}

The notion of unconditional counterfactuals as explanations for automated decisions was proposed by Watcher et al. \cite{Watcher}. They defined "good" explanations as those that can 1) \emph{inform and help understand why the decision was made}. However, just understanding is not enough, as they also \emph{provide grounds to contest the decision}. Since the aim of a good explanation is to inform us on how to help ourselves, they should \emph{help understand what can be done to get the desired outcome}. 

The use of \emph{Unconditional Counterfactuals(Cfs)} was proposed as a solution to this problem of obtaining "good" explanations. They provide meaningful explanations for individual decisions with regards to \emph{understanding why the decision was made}, \emph{how to contest it}, and \emph{advice on how the individual can change their situation to obtain the desired outcome}.

Counterfactual explanations come with a few properties. Namely, they are intentionally restrictive. They do not require an understanding of the internal logic employed by the decision-making system. In addition to this, they are also privacy-focused. They are designed to provide the minimum amount of information that is required to alter the decision. In addition to this, they are also used to identify if algorithms are fair with respect to protected groups and populations, e.g., The job will be given if the sex of the applicant is Male.

One of the methods to generate counterfactuals was through a technique called \emph{adversarial perturbations}. It is designed to generate counterfactuals, as highlighted by Watcher et al. \cite{Watcher}. The subsequent counterfactuals are synthetic data points that are classified differently. The work of Watcher et al.  \cite{Watcher} highlighted how a majority of such approaches do so by making small changes to many variables instead of providing sparse human interpretable solutions that modify only a few variables. 

Watcher et al.  \cite{Watcher} approached the problem as an optimization problem. The goal is to find the counterfactual that is the least distance away from the original data point while maximizing the probability of achieving the desired outcome. The lack of human interpretability of the earlier approaches was addressed through the distance function, which is used to decide the closer synthetic points. 

There are several advantages to using \emph{counterfactuals} as explanations. Firstly, they bypass the substantial challenge of explaining the inner workings of complex machine learning (ML) and artificial intelligence (AI) based systems. Not all sections of the population will find the explanations equally meaningful. Hence counterfactuals provide information that is both easily digestible and useful for understanding. It is able to explain some of the rationales of automated decisions while avoiding the pitfalls of interpretability.

In addition to this, they satisfy the requirements of a "good" explanation. Firstly, they inform us about \emph{the reason behind a decision}. Counterfactuals do so in a manner that bypasses the substantial challenges of explaining the internal workings of complex ML systems. They also provide information to us that is both easily digestible and potentially useful. 

Secondly, they provide grounds for \emph{contesting /challenging decisions}. They do so by providing information on the external factors that contribute to a decision. Hence they highlight the information required to exercise the right to contest all without explaining the inner workings of the decision-making algorithm. 

Lastly, counterfactuals inform us on \emph{how altering future behavior might result in the desired outcome}. It makes use of explanations to indicate what could be changed to receive the desired result in the future. By providing information about "close possible worlds" which result in different decisions, individuals can understand what factors need to be altered to get the desired outcome. That being said, unanticipated dependencies between altered attributes and other variables may undermine the utility of counterfactuals as a guide for future behavior.

\subsection{Constrained Most Probable Explanation(CMPE)}

There have been various algorithms to address the most probable explanation problem. The work by Rouhani et al. \cite{Gog_ref_CMPE} came up with a novel algorithm that solves the \emph{constrained most probable explanation (CMPE)} problem: given a set of random variables and two possibly identical models, it seeks to find the most likely assignment of variables with one of the models such that its probability of assignment of the second model is smaller than a real number.  One of the novel applications of CMPE is that it gives the most likely changes to be made to a test sample to achieve a desired outcome by a classifier. This is analogous to the \emph{counterfactuals} as described by Watcher et al. \cite{Watcher}.

One of the advantages of the CMPE algorithm is that several explanation and prediction tasks can be reduced to CMPE, e.g., the nearest assignment problem \cite{Gog_ref_NAP}. It does so by combining existing algorithms for partitioning graphs with algorithms developed for tackling the multiple choice knapsack (MCKS) problem. Specifically by reducing the CMPE to an instance of a bounded MCKP, which is a weakly NP-hard problem compared to a strongly NP-hard problem. This is the combined primal graph has multiple connected components each of which having a fixed number of variables ($k$). There are existing approaches for tacking the MCKP task with guarantees that are used to solve the CMPE task. It was observed through experiments that most CMPE problems are easy when $k$ is small. Hard instances arise when $k$ is close to the unconstrained maximum or the graph's model parameters are extreme. These hard instances benefit from using a large value of k, while easy instances do not. Rouhani et al. \cite{Gog_ref_CMPE} attributed this to the \emph{exploration vs. exploitation} trade-off. When k is large, it explores more states. For easy instances, there are an abundant number of optimal solutions. When compared to hard problems, there are relatively few optimal states to search for. Hence having a large k means the algorithms spend more time on each state. It was shown how the CMPE approach labelled \emph{ANYTIME-CMPE} was superior to a state of the art mixed interger liner programming solver (MILP)

The work as described above yield a lower bound for the CMPE. Rahman et al. \cite{Gog2} worked towards proposing approaches for efficiently computing upper bounds for CMPE. They had two approaches: either relax the objective, the constraint, or both. The first approach relaxes the global constraints based on \emph{Lagrangian relaxation} \cite{Gog_ref_Lagrange} to yield a Most Probable Explanation task. The second approach is based on \emph{Lagrangian decomposition} and results in a Multiple Choice Knapsack Problem(MCKP). Their work explores the novel application of making classifiers change their decisions by minimally altering the example. This can be thought of as a form of \emph{counterfactual} \cite{Watcher}.

The first approach is to relax the global constraints. Making use of Lagrange relaxation reduces the task to an unconstrained MPE task. This approach gives the best results when there are multiple connected components. The second approach involved reducing the CMPE task to a MCKP. The MCKP is then solved using specialized algorithms. Each approach has its advantages and disadvantages. If the primal graph has a large number of disconnected components, the second approach yields better quality bounds than the former. However, if the MPE is tractable, then MPE-based bounds are superior in terms of quality than MCKP-based bounds.

\subsection{Counterfactuals in Explainable AI}
We have explored the novel, \emph{model-agnostic} algorithm LIME. While it does have its benefits, LIME also has its drawbacks. Through the work by Lundberg and Lee \cite{SHAP},  it was discovered that the LIME approach, as it was introduced, did not satisfy the properties of \emph{local accuracy} or \emph{consistency} as it did not make use of the unique solution that offers such a guarantee-SHAP values. This was explored further through the work by White et al. \cite{CLEAR}. Their work highlighted the shortcomings of LIME, specifically its failure to produce meaningful explanations locally. 

White et al. \cite{CLEAR} took an approach of measuring the \emph{fidelity of the explanation model to the underlying classifier} to evaluate how meaningful the explanations actually are. To do this, they made use of \emph{counterfactuals} as described by Watcher et al.

LIME makes use of \emph{local explanations} to explain the model as a whole. Similarly to this, the approach introduced by White et al. \cite{CLEAR} called \emph{Counterfactual Local Explanations for Any Classifier} (CLEAR) also leverages \emph{local} explanations to provide explanations \emph{globally}. CLEAR makes use of counterfactual analysis to improve on the shortcomings of LIME which neither measures its own \emph{fidelity} or generate counterfactuals. This use of counterfactuals results in more meaningful explanations with improved local \emph{fidelity} (by an average of 40 \%), resulting in improved performance; on real-world datasets like \emph{adult}, \emph{credit}, the regressions of CLEAR are better than that of LIME.

LIME samples synthetic data instances locally for its \emph{explanation model}. These are generated using perturbations. Making use of these synthetic observations, LIME produces a \emph{locally} weighted regression whose coefficients are used to explain the model's predictions. However, in LIME, the \emph{local} explanation model is not required to correctly predict the label. In addition to this, the synthetic data instances sampled to be from the local neighborhood might end up creating an imbalanced synthetic dataset of local instances which reduces the \emph{fidelity}. Counterfactuals are used for the sampling of the synthetic local instances. Making use of counterfactuals, synthetic instances that are equally distributed across classes can be generated. The resulting balanced dataset produced regressions with improved \emph{fidelity} compared to using imbalanced datasets.

\begin{table}[htbp] 
\begin{tabular}{||c c c c c c||} 
 \hline
  & Adult & Iris & Pima & Credit & Breast\\ [0.5ex] 
 \hline\hline
 CLEAR with counterfactuals & 80\% $\pm$ 0.8  & 99.8\% $\pm$ 0.1 &  77\% $\pm$ 0.8 & 55\% $\pm$ 1.7 & 81\% $\pm$ 1.3\\ 
 \hline
 CLEAR without counterfactuals & 80\% $\pm$ 0.9 & 80\% $\pm$ 1.0 &  57\% $\pm$ 0.8 & 39\% $\pm$ 1.3 & 54\% $\pm$ 1.1\\ 
 \hline
LIME & 26\% $\pm$ 0.6 & 30\% $\pm$ 0.3 &  20\% $\pm$ 0.4 & 12\% $\pm$ 0.5 & 14\% $\pm$ 0.3\\  [1ex] 
 \hline
\end{tabular}
\caption{Performance of CLEAR with counterfactuals vs. CLEAR without counterfactuals vs. LIME}
\end{table}

\subsection{Counterfactuals and the Human Thought Process}

Counterfactuals are being increasingly used to improve the process of getting to trust models. For humans, an interpretable model allows one to 1) mentally simulate some aspects of the system and understand the causes of the decision-making and 2) enables the user to consider contrastive explanations and \emph{counterfactual analysis} (why was a decision made and if the feature had changed, what would the output be).

Given that the ultimate goal of machine learning and artificial intelligence is to improve the lives of human beings, rather than explain the decision, it should provide a means to help realize the decision. In the paper by Byrne \cite{Byrne19}, this is explored further. Similar to how humans think, the paper highlighted how counterfactuals are thought of and how humans go about the formation and reasoning of the same. 

In the work of Byrne \cite{Byrne19}, it was also highlighted how counterfactuals can be used to identify cause-effect relationships, thereby improving the interpretability of the model. One of the advantages of using counterfactuals is that one can use them to get an idea of what would have to change to achieve the desired outcome. Byrne \cite{Byrne19} highlighted areas that are identified as ripe for change, there is a pattern in what humans choose, namely exceptions, controllability, and action compared to failure to act: Tend to create counterfactuals on the act vs. the failure to act, Recent Events, Probability - Alter events that are feasible.

\subsection{Recourse in Counterfactual Explanations}

The idea of \emph{counterfactuals} as a source of explanation was first introduced by Watcher et al. \cite{Watcher}. As explained earlier, it highlights the minimum change necessary to get the desired outcome. Unfortunately, it doesn't take into account the nature of relationships that govern the world we live in, e.g., If you were a male, you would have been approved for the loan. As a result, counterfactual explanations are offered that offer solutions that we can never hope to achieve.

Ustun et al. \cite{Rec} explored this problem of \emph{algorithmic recourse}. \emph{Recourse} is defined as the ability of an individual to obtain the desired outcome from a given model. It identifies the lack of autonomy as a source of unfairness and seeks recourse as the solution. They saw that providing recourse to all individuals is only guaranteed in some of the existing research. This makes sense as explanations might be offered that make extremely difficult to even impossible suggestions, e.g., raise your salary by 300\% or change your gender or your ethnicity.

\begin{figure}[htp]
    \centering
    \includegraphics[width=9cm]{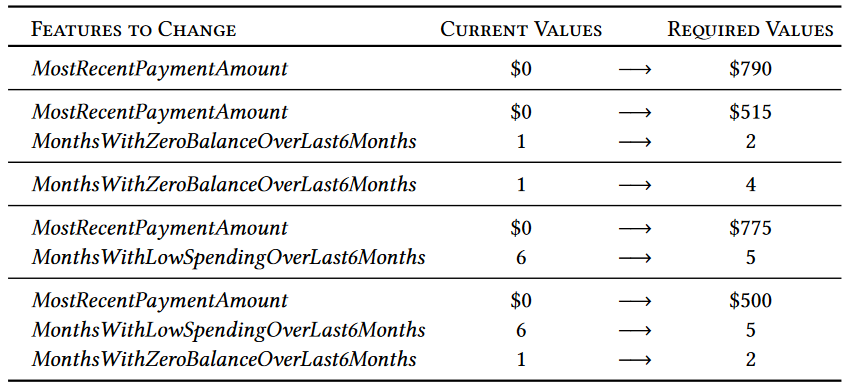}
    \caption{An example of the flipsets resulting from the Actionable recourse method}
    \label{fig:Recourse}
\end{figure} The work by Ustun et al. \cite{Rec} aimed to explore the feasibility of \emph{universal recourse}. This means that all individuals have a mechanism to get the desired outcome. This was done through 1) exploring how the choice of features used affects recourse, 2) evaluating how out models deployed in adversarially distributed populations affect the feasibility of recourse. 

There were several benefits to their approach a) it is designed to ensure recourse without interfering in the model development. In addition to this, b) they explore how recourse changes across individuals and demographic groups. Lastly, 3) their methods provide a way of generating a set of diverse alternate solutions. They show all the possible changes that one needs to make to get the desired outcome. 

They created a tool and a procedure to evaluate the feasibility and difficulty of recourse in decisions. This is extremely important as a model that does not guarantee recourse is not fair for the individual. 

Their tools also offer a list of diverse counterfactual suggestions to choose from. This is similar to how we as humans would operate, e.g., if your salary rose by 15\%, you would be approved for the loan; if your savings show \$100,000 dollars, then you would be approved for the loan. They refer to these as flipsets. This method generates a list of alternate actionable changes that can be made to obtain the desired outcome. 

\begin{figure}[htp]
    \centering
    \includegraphics[width=9cm]{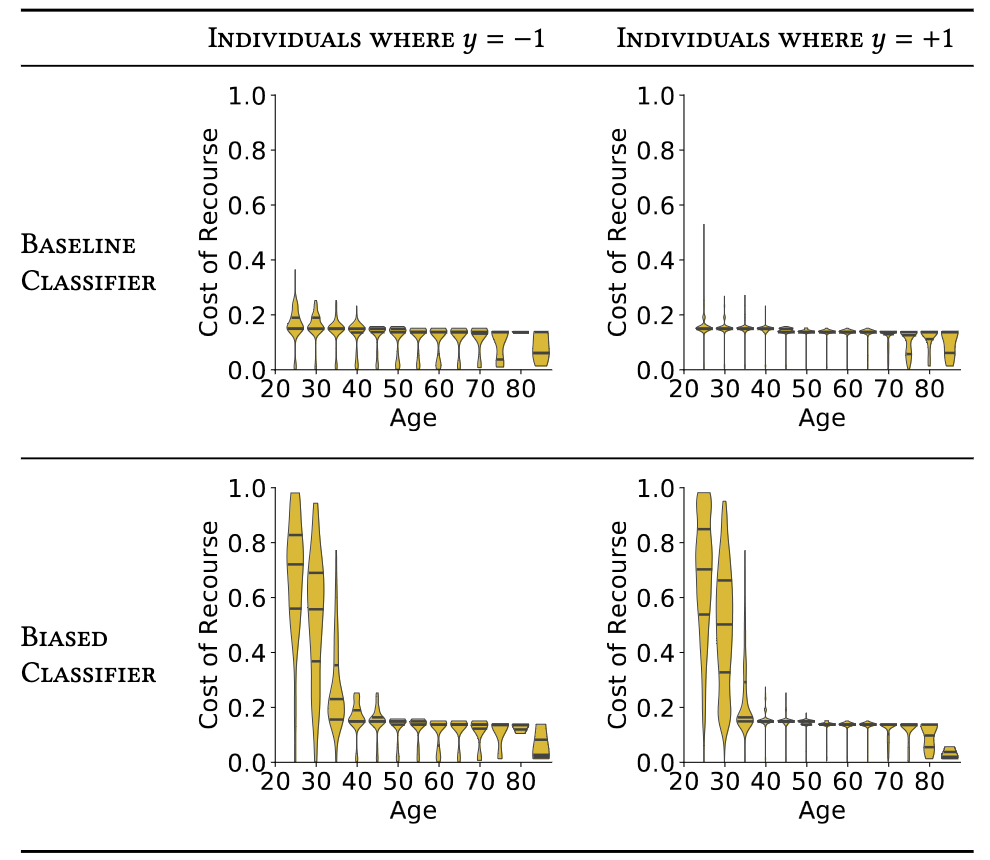}
    \caption{Distributions of the cost of recourse in the target population. It shows that when models are deployed on out-of-sample data, it incorrectly attributes cost in terms of recourse. The biased classifier was trained on data devoid of any samples below the age of 35. Hence the resulting cost of recourse for such individuals is higher }
    \label{fig:Rec}
\end{figure}
While these tools do have their benefits, they are only applicable to linear classification models like linear classification models, e.g., logistic regression models, linear SVMs, and linearizable rule-based models. Their models provided a universal \emph{recourse} guarantee. It stated that a linear classifier provides \emph{recourse} to all individuals if it only uses actionable features and does not predict a single class. The converse of it is also true; a classifier will deny \emph{recourse} if it uses immutable features exclusively or predict a single class consistently. This highlighted the dependence of \emph{recourse} on the actionability/immutability of features, e.g., if you were a male, you would be given the loan. Since one cannot change genders, they are denied \emph{recourse}. 

Their work also highlighted how \emph{recourse} is not guaranteed if classifiers use features that are immutable or conditionally mutable. They also highlighted how features being bounded or unbounded affects \emph{recourse}. If all features are unbounded, then a linear classifier with at least one actionable feature provides \emph{recourse} to all individuals. However, if all features are bounded, then a linear classifier with at least one immutable feature may deny \emph{recourse} to some individuals.

Another cause for denial of \emph{recourse} was identified as out-of-sample deployment. This happens because models are trained on populations that are different from the populations that it is deployed against. Ustun et al. \cite{Rec} identified how this out-of-sample deployment results in the model incorrectly attributing higher costs for specific actions, thereby making \emph{recourse} increasingly difficult to even impossible. 

Another use of the methods from Ustun et al. \cite{Rec} is evaluating the disparities in \emph{recourse} across populations. They identified this by seeing how the difficulty of \emph{recourse} changes across individuals from different sub-populations, e.g., gender and ethnicity.

\subsection{Model Agnostic Counterfactual Explanations (MACE)}

We have spoken about the requirement of a good explanation  1) it should answer why it predicted a particular output, and  2) it should inform on what changes should be made to obtain the desired output. Karimi et al. \cite{MACE} focused on answering the latter. Their solution to this was through the use of counterfactual explanations. 

With regard to \emph{counterfactuals as a solution for explanations}, several works have been proposed which make use of optimization-based methods to generate the nearest counterfactual explanation. The nearest counterfactual explanation refers to the counterfactual that, for the given model, 1) results in a different decision and 2) is a result of minimal changes made to the original factual instance. Watcher et al. proposed generating counterfactuals that were restricted to differentiable models. The work by Ustun et al., while evaluating recourse, was restricted to linear models. 

In contrast, the work of Karimi et al. made use of the knowledge of counterfactual reasoning to come up with a \emph{model-agnostic} algorithm to generate such counterfactual explanations that achieve the desired outcome. Their novel algorithm called  \emph{Model Agnostic Counterfactual Explanations} (MACE)  solves a series of satisfiability problems where the objective and constraints are expressed as logical formulae. This MACE approach comes with the benefit of being \emph{model-agnostic},  data type agnostic, and distance agnostic. In addition to this, it is able to generate plausible and diverse counterfactuals for any sample and therefore has 100\% coverage. Similar to how Ustun et al. computed the counterfactual in the form of flip sets with minimal cost,  \emph{Model Agnostic Counterfactual Explanations} (MACE) Karimi achieves the same. The resulting counterfactuals, while being diverse and plausible, are also closer than those through other methods. 
\begin{figure}[htp]
    \centering
    \includegraphics[width=11cm]{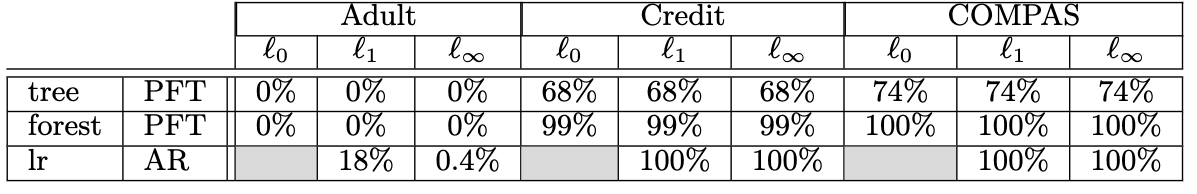}
    \caption{Coverage 500 factual samples. MACE and MO (Minimum Observable approach) has  100\% coverage always, by
definition and by design, respectively. PFT is the modified version of the Feature Tweaking algorithm }
    \label{fig:MACE}
\end{figure}

In MACE, the objective, as well as the constraints, are expressed as logical formulae. This allows for easily supporting additional plausibility constraints. It was seen that compared to existing approaches, MACE outperforms them on real-world datasets. It has the advantage of achieving 100\% coverage which was not achieved by Ustun et al.. This flexibility allows extending plausibility constraints to different features like actionable, immutable, or conditionally mutable, e.g., the feature age can only increase and not decrease, and the feature- sex cannot be changed. The work of Ustun et al. gave a set of diverse counterfactual solutions, also referred to as flip sets. MACE also produces its own flip sets. It is done through the use of diversity constraints.

On experimenting with existing such as Minimum Observable (MO) \cite{MACE_ref_1} , Feature Tweaking (FT) \cite{MACE_ref_2}, and Actionable Recourse (AR) \cite{Rec}, MACE outperformed them. With respect to coverage, MACE consistently achieves 100\% coverage. The same cannot be said for  Actionable Recourse. Additionally, the resulting counterfactuals are significantly closer than those produced by other methods. As a result, they require significantly less effort in order to be achieved.

Overall, MACE presented an approach to generating counterfactual explanations in the context of consequential decisions. By combining formulae for the model, distance, plausibility, and diversity, they demonstrated how MACE not only achieves 100\% coverage but also generates counterfactuals at closer distances when compared to optimization-based approaches. Similar to the work of Ustun et al., it can evaluate the disparity of recourse across protected attributes like gender and ethnicity.
\begin{figure}[htp]
    \centering
    \includegraphics[width=11cm]{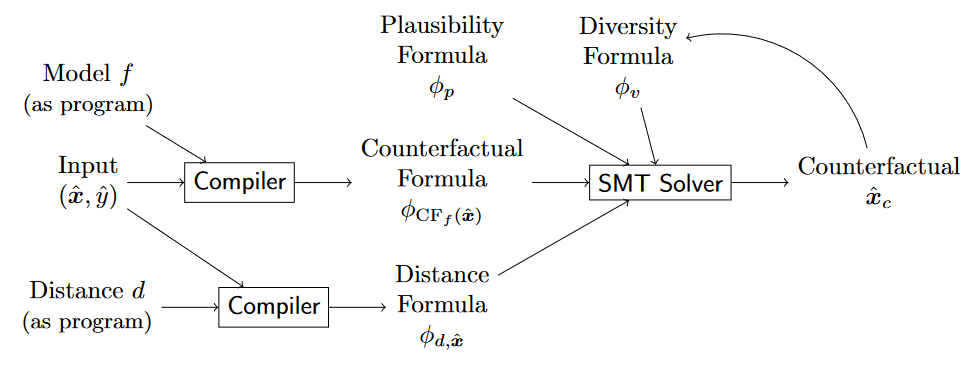}
    \caption{Architectural overview of the Model Agnostic Counterfactual Explanations method}
    \label{fig:MACE}
\end{figure}

\subsection{Causality in XAI}

Pearl \cite{Pearl} highlighted how most of the use cases deal with problems that are causal and not associational, e.g., what is the cause of the illness? They are causal as they cannot be computed from the data alone or from the distributions governing the data. Statistical analysis focuses on finding associations. Causal analysis, on the other hand, aims to infer not only beliefs but changing beliefs. Essentially, this highlights how causal and associational concepts do not mix. The joint probability distribution does not tell us how it would be different if external conditions change. This information is obtained through causal assumptions that identify relationships that remain invariant to external change.

Associational concepts are relations that can be defined in terms of a joint probability distribution; on the other hand, causal concepts are relations that cannot be defined from a distribution alone. Every claim invoking a causal concept relies on the premise that invokes concepts; it cannot be inferred from; or even be defined in terms of statistical associations alone. \emph{Behind every causal conclusion lies some causal assumptions that are untested in observational studies}. As a result of this distinction between associational and causal concepts, the need for a new notation to express causal analysis arose. This was as Pearl \cite{Pearl} highlighted how probability distributions are insufficient, e.g., they cannot identify if a symptom causes a disease or if a disease causes a symptom. So the tools used in statistics are insufficient for expressing causal relations.  

Pearl \cite{Pearl} came up with a general theory for causality based on the Structural Causal Model (SCM)\cite{Pearl_ref_1, Pearl_ref_2}, in which the world was expressed in the form of structural equations that model the causal dependencies of the world. 

Causal assumptions were not encoded in links but missing links. Each assumption in itself cannot be tested(since causal analysis requires making causal assumptions that are not testable unless in experimental settings); however, the sum total of all causal assumptions has testable applications. It is the independence identified through \emph{d-separation} \cite{Pearl_ref_3} that allows the assumptions in structural equation models to confront the scrutiny of non-experimental data. 

The problem that Pearl focused on is if we make changes in the form of interventions, can the post-intervention model be accurately described by the pre-intervention model, i.e., can we accurately predict the effects of our intervention? Pearl focused on the methods of modeling causality and also highlighted the conditions that would allow the prediction of the post interventions effects. Pearl made use of graphical criteria known as d-separation \cite{Pearl_ref_3}. 
\begin{figure}[htp]
    \centering
    \includegraphics[width=11cm]{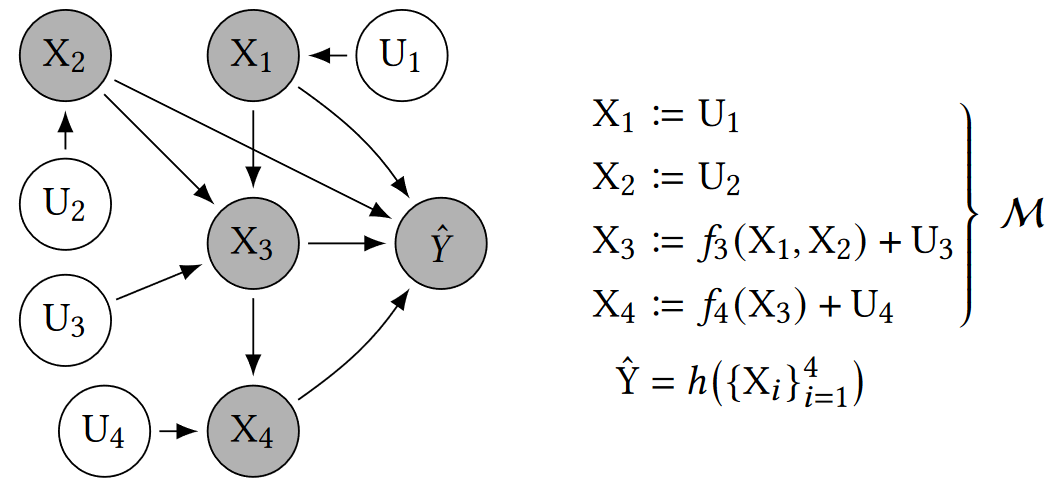}
    \caption{This is an example of a structural causal model}
    \label{fig:MINT}
\end{figure}

\subsection{Counterfactuals as Interventions}

We have explored various works on counterfactuals as a solution to the explanation problem. Unfortunately, they all share a similar trait in that: their objective is to find the nearest counterfactual explanation. As mentioned earlier, the nearest counterfactual explanation has the property that it is the result of the minimum shift from the original instance. The assumption is that the nearest counterfactual in terms of distance is also the most easily attainable in terms of effort or cost. In other words, nearest counterfactual explanations inform an individual where to go but not how to get there. The work of Karimi et al. relied on causal reasoning to highlight the shortcomings of using the nearest counterfactual explanation. They instead proposed shifting the focus from finding the nearest counterfactual explanations to one finding the minimum set of interventions that result in the desired outcome.

The assumption is that the set of actions resulting in the desired outcome would directly follow from the nearest counterfactual explanation. This, as explored in the work by Karimi et al. \cite{MACE_R}, was not true. This, as explored by the authors, is due to the lack of consideration of the causal relationships that govern the world. As seen by them, the actions derived directly from counterfactual explanations may ask for too much effort from the individual or may not even result in the desired outcome.

To tackle this problem, Karimi et al. \cite{MACE_R} reformulate the \emph{recourse} problem. Using causal reasoning, they incorporate knowledge of causal dependencies into the process of recommending recourse actions. If followed through, these actions would result in a counterfactual instance that favorably obtains the desired outcome. Hence they proposed a method of recourse through minimal interventions. The solutions, as a result, inform us how to act in addition to understanding. 

The proposed methods seek a minimal-cost set of actions in the form of structural interventions. These structural interventions result in structural counterfactuals, which are less costly in terms of effort compared to those obtained through the nearest counterfactual. In the german dataset, there is a 42\% reduction in effort. This is due to the fact that nearest counterfactual explanations fail to take advantage of the causal relations between features.

Karimi et al. \cite{MACE_R} identified different forms of interventions, each with its own advantages and disadvantages. Hard interventions sever any relation between variables. There are helpful when we wish to isolate the causal effect of an action, e.g., the effect of smoke on the lungs. Soft interventions, on the other hand, do not sever the relationships but instead, take advantage of them.

Another advantage of this approach is the fact that features that were once not directly actionable can now be acted upon, e.g., \emph{credit score} cannot be altered directly, but the features of \emph{regular payment}, and \emph{salary} can indirectly affect credit score. By taking advantage of these relations, intervening upon ancestors to cause a down steam effect is possible.

\section{Discussion}

We have covered various approaches to explaining/interpreting the decisions made by various models. We covered linear-based models and tree-based models like decision trees. Both are intrinsically explainable models. Since their focus is on explainability over performance, as long as the underlying data is simple enough (not too complex), these are appropriate.

On the other hand, we have seen a plethora of complex and highly accurate models. Their explainability is limited by their complexity, e.g., deep learning approaches. In this case, post hoc explainability approaches would be appropriate. Some examples are LIME , KernelSHAP. There are many instances where One might need to work with different data types. LIME, SHAP being model agnostic, can be used in these cases.

There might be instances where we are more focused on individual/\emph{local explainability}, e.g., a bank manager, as well as the customer, might want to know why they were rejected for a loan by the application. In these cases, local explanations are more than sufficient as long as global understanding is not required. Counterfactuals are an excellent example of these. They can be used to identify cause-effect relationships and also have the added benefit of making suggestions to help obtain the desired outcome. On the other hand, \emph{global explainability} approaches are tasked with understanding the underlying model's logic as a whole. In those cases, approaches like FOLD and FOLD-R++ (for tabular-based data) will work. There are instances where we might want answers that are intuitive. The FOLD family of algorithms provides explanations in the form of \emph{defaults and exceptions} which can be used in such cases.

There are many instances where counterfactual suggestions might be challenging to achieve, if not impossible. Ustun et al. made an approach called Actionable Recourse to evaluate the recourse provided by counterfactual explanations. In addition to this, their approach also allows the generation of diverse counterfactuals to choose from. However, this is limited to linear settings. The MACE algorithm outperforms Actionable Recourse with universal coverage. Unfortunately, sometimes counterfactuals suggested might not be accurate, as in they are a lot harder to achieve or are counter-intuitive. These are usually in cases when features are causally related. The work by Watcher et al. was limited to making some assumptions- the features were independent. This is, however, different in the real world. Watcher et al. did not focus on causal dependence as it was beyond the scope of their work. By using the frameworks developed by Pearl, Karimi created the MINT approach, which provides explanations as counterfactual interventions while taking causal relationships into account. 

Shakerin highlighted how the FOLD algorithm that models the human thought process could benefit greatly by incorporating the knowledge of counterfactuals. Shakerin proposes  Craig Interpolants \cite{FOLD_ref_1} as an approach to try to achieve this.

\section{Conclusion}

Throughout this survey, we have explored the various approaches towards making machine learning models more comprehensible to human beings. We have covered the various categories of approaches from \emph{intrinsically explainable} to \emph{post hoc} explanations. We have discussed the differences between \emph{model agnostic} and \emph{model specific} approaches as well as \emph{global and local approaches}. In addition to this, there are also approaches for dealing with different types of data such as tabular, text and images.  

After this, we have explored some of the popular algorithms for providing explanations. We started with LIME and explored its advantages. Going even further, we were able to appreciate the advantages provided by SHAP values which unified six different approaches of explanation models. The resulting explanations are more in line with human intuition. After exploring SHAP values, we identified explanation methods that use \emph{common sense reasoning} and \emph{default theory}. We can appreciate the human intuitive explanations provided by FOLD, FOLD-R and FOLD-R++ which are in the form of \emph{default rules}. Unfortunately, explanations with limited utility are meaningless to us. This was explored through the CLEAR algorithm which improved on the LIME approach using counterfactuals. Counterfactuals are a form of explanation that explain decisions locally. Being introduced as a viable form of explanation, they were used to provide individual explanations in the form of a response to a decision. 

Unfortunately, it was later seen that when considering counterfactuals as a form of explanation, recourse should be taken into account. Seeing the impact of recourse and how explanations ultimately need to help us as people, we have explored the difficulty in realizing counterfactuals. In this sense, we have identified the need to include causality in our approaches.

My intention is to incorporate causality into common sense reasoning to produce more human-intuitive counterfactual explanations both as a means to provide explanations as well as suggest possible courses of action.

%% If your work has an appendix, this is the place to put it.
\appendix

\bibliographystyle{ACM-Reference-Format}
\bibliography{sample_base_2}

%%
%% If your work has an appendix, this is the place to put it.
\appendix

\end{document}